\documentclass{article}
\pdfoutput=1
\usepackage[preprint]{neurips_2024}

\usepackage{enumitem}
\usepackage{amsmath,amsfonts}
\usepackage{algorithmic}
\usepackage{graphicx}
\usepackage{textcomp}
\usepackage{listings}
\usepackage{tabularx} % Add this in the preamble
\usepackage[table]{xcolor}
\usepackage{pifont} 
\usepackage{subcaption}
\usepackage{balance}
\usepackage{wrapfig}
\usepackage{booktabs}
\usepackage[utf8]{inputenc}
\usepackage[normalem]{ulem}
\usepackage{url}
\usepackage{xspace}

\newcommand{\rmodel}{{LangCache-Embed}\xspace}

\usepackage{pgfplots}
\usepackage{pgfplotstable}
\usepackage{amsmath}
\pgfplotsset{compat=1.18}
\usepackage{graphicx} % Required for inserting images
\usepackage{adjustbox}
\usepackage{subcaption}
\usepgfplotslibrary{groupplots} % Required for subplots
\usepackage{listings}

\lstdefinestyle{customprompt}{
  backgroundcolor=\color{gray!10},   
  basicstyle=\ttfamily\small,
  breaklines=true,                 
  captionpos=b,                    
  numbers=none,                    
  frame=single,                    
  rulecolor=\color{black},
  showstringspaces=false,
}

\usepackage{filecontents}

\title{Advancing Semantic Caching for LLMs with Domain-Specific Embeddings and Synthetic Data}

\author{%
  Waris Gill$^{1,2}$ \quad Justin Cechmanek$^{1}$ \quad Tyler Hutcherson$^{1}$ \quad Srijith Rajamohan$^{1}$\\[1ex]
  \textbf{Jen Agarwal}$^{1}$ \quad \textbf{Muhammad Ali Gulzar}$^{2}$ \quad \textbf{Manvinder Singh}$^{1}$ \quad \textbf{Benoit Dion}$^{1}$\\[1ex]
   $^{1}$Redis, USA
   \quad
  $^{2}$Virginia Tech\\[1ex]
  \texttt{\{waris.gill, justin.cechmanek, tyler.hutcherson, srijith.rajamohan,}\\
  \texttt{jen.agarwal, manvinder.singh, benoit.dion\}@redis.com}\\
  \texttt{waris@vt.edu, gulzar@cs.vt.edu}
}

\begin{document}

\maketitle
% \documentclass{article}
% \usepackage{graphicx} % Required for inserting images
% \input{macros}

% % \title{Accelerating LangCache with Fine-Tuned Embeddings: Towards Optimal Cache Hits}
% % or

% \title{Rough Title: On Advancing Semantic Caching with Specialized Embedding Models}

% % we can add `LangCache` keyword in the title as well.

% \begin{document}
% \maketitle

\begin{abstract}
  This report investigates enhancing semantic caching effectiveness by employing specialized, fine-tuned embedding models. Semantic caching relies on embedding similarity rather than exact key matching, presenting unique challenges in balancing  precision, query latency, and computational efficiency. We propose leveraging smaller, domain-specific embedding models, fine-tuned with targeted real-world and synthetically generated datasets. Our empirical evaluations demonstrate that compact embedding models fine-tuned for just one epoch on specialized datasets significantly surpass both state-of-the-art open-source and proprietary alternatives in precision and recall. Moreover, we introduce a novel synthetic data generation pipeline for the semantic cache that mitigates the challenge of limited domain-specific annotated data, further boosting embedding performance. Our approach effectively balances computational overhead and accuracy, establishing a viable and efficient strategy for practical semantic caching implementations.

\end{abstract}

%test

\section{Introduction}

% \the\textwidth

Large language models (LLMs) are rapidly becoming integral components of modern applications, significantly influencing everyday tasks. These models consist of billions of neurons and require substantial computational infrastructure to operate effectively. Each user query to an LLM involves billions of floating-point operations to generate an appropriate response \cite{OPTQ}.

In practice, users often issue repeated queries to online services. Prior research ~\cite{lempel2003predictive, xie2002locality, markatos2001caching} indicates that approximately 33\% of queries submitted to web search engines are repeated. A similar phenomenon occurs with LLM-based services~\cite{gill2025meancacheusercentricsemanticcaching}. Recognizing this, researchers have proposed implementing a \textit{semantic cache} to efficiently handle duplicate queries directed at LLMs~\cite{gptcache, gill2025meancacheusercentricsemanticcaching, zhu2023towards}. Unlike traditional key-value caches, a semantic cache does not rely on exact key matching. Instead, it declares a \textit{cache hit} if the similarity between the embeddings generated from two textual inputs surpass a predefined cosine similarity threshold ~\cite{gptcache, gill2025meancacheusercentricsemanticcaching}.

A semantic cache generally comprises two essential components: an embedding model, which computes query embeddings, and a vector database, which stores these embeddings and retrieves cached responses by matching embeddings of repeated queries. A variety of embedding models are available, ranging from open-source to proprietary closed-source models, both demonstrating state-of-the-art (SOTA) performance on benchmarks such as MTEB~\cite{muennighoff-etal-2023-mteb}. An important consideration thus arises: which embedding model is optimal for semantic caching? Both open-source and closed-source options have their respective advantages and disadvantages. Open-source models are freely accessible, allowing users to maintain data privacy by running models on their own infrastructure. However, these models often contain billions of parameters, as exemplified by Alibaba-NLP/gte-Qwen2-7B-instruct (a 7-billion-parameter embedding model~\cite{li2023towards, AlibabaN57:online}), making them computationally intensive and less suitable for efficient semantic caching. Conversely, closed-source embedding models offered via managed services are costly, may raise data privacy concerns due to external data handling, and introduce network latency, potentially hindering the performance of semantic caches.

Ideally, an embedding model used in semantic caching should be lightweight and computationally efficient. Unfortunately, smaller models typically lag behind their larger counterparts in performance. Nevertheless, recent studies have demonstrated that smaller, task-specific fine-tuned models can surpass the performance of significantly larger models~\cite{ouyang2022training, penedo2024refinedweb, du2024compositional}. Motivated by this insight, we fine-tuned smaller embedding models on domain-specific datasets (medical and Quora~\cite{kaggleQuestionPairs} datasets). Remarkably, our experiments revealed that fine-tuning for just one epoch enabled these compact models to outperform both state-of-the-art closed-source and open-source models.

However, fine-tuning these smaller models requires high-quality datasets, which may not always be readily available. To address this limitation, we developed a unique synthetic data generation pipeline for the semantic cache. This pipeline leverages existing datasets from diverse domains to produce targeted synthetic data tailored specifically for different applications and domains. Our evaluation demonstrates that synthetic data significantly enhances the smaller embedding model performance for semantic caching, achieving precision improvements of 9\% compared to its non-finetuned base model. Moreover, when tested on real-world medical queries, the model fine-tuned on medical synthetic data achieves performance approaching or even matching that of leading open-source and closed-source models, surpassing OpenAI's embedding model by 2\% in precision.

\section{Methodology}

Embedding models lie at the heart of semantic caching. Their role is to produce high-quality, high-dimensional representations that encapsulate the semantic information in text. The better these models are at distinguishing subtle nuances in meaning, the more effective the cache becomes at reusing relevant results. State-of-the-art embedding models, whether open-source (such as Alibaba-NLP/gte-Qwen2-7B-instruct) ~\cite{li2023towards, AlibabaN57:online, zhang2024mgte}  or closed-source~\cite{AmazonTi29:online, EmbedThe12:online,Vectorem3:online,lee2024gecko, lee2025gemini, neelakantan2022text} , have shown remarkable effectiveness in tasks like sentence similarity, text classification, and semantic retrieval. In semantic caching, however, the overarching challenge is balancing performance with computational and operational constraints. While larger models often excel in capturing fine-grained nuances, they are costly to run repeatedly at scale. Conversely, more efficient, smaller models may falter on challenging queries, leading to suboptimal retrieval and degraded cache performance.

A key shortcoming of baseline (out-of-the-box) embedding models is that they are typically trained on broad, general-purpose corpora. When faced with domain-specific queries (e.g., finance, medical, or legal), these models may struggle to capture the nuanced relationships unique to that domain. This leads to lower performance, especially for duplicate queries that differ slightly in wording but share the same semantic content. For instance, in the medical domain, the queries ``myocardial infarction treatment" and ``how to treat a heart attack" may not be recognized as semantically equivalent by a general-purpose embedding model. Such limitations mean that, in practice, relying solely on these larger, general-purpose embedding models can either incur excessive compute overhead or fail to retrieve sufficiently accurate cached results for highly specialized queries. Consequently, there is a growing need for fine-tuned, domain-adapted embedding models, particularly compact ones, that can provide robust semantic representations while remaining efficient to deploy in production. To address the challenges outlined above, our methodology comprises three main components: (1) Model Selection and Training, (2) Domain-Specific Fine-Tuning, and (3) Synthetic Data Generation.

% We begin by choosing compact, publicly available embedding model, ModernBERT~\cite{} containing  (e.g., 149 million parameters). ModernBERT is recent encoder only transfomer that outperforms similar existing transformers such as Bert, NomicBert, RoBERTa. Note that our insights are equally applicable with other smaller models such as Mini-LLM, MPNet and Albert. We choose ModernBERT based on its efficieny and performance.  Models like ModernBERT advantageous for semantic caching because of their reduced computational footprint and faster inference speeds.
We begin by selecting a compact, publicly available embedding model, ModernBERT~\cite{warner2024smarter}, which contains approximately 149 million parameters. ModernBERT is a recent encoder-only transformer that outperforms comparable models such as BERT~\cite{devlin-etal-2019-bert}, NomicBERT~\cite{nussbaum2025nomic}, and RoBERTa~\cite{liu2019robertarobustlyoptimizedbert}. While our insights are broadly applicable to other smaller models like MiniLM~\cite{wang2020minilm}, MPNet~\cite{song2020mpnet}, and ALBERT~\cite{Lan2020ALBERT}, we choose ModernBERT for its strong balance of efficiency and performance. Models like ModernBERT are particularly advantageous for semantic caching due to their reduced computational footprint and faster inference times.

To fine-tune the ModernBERT embeddings for domain-specific queries, we use the online contrastive loss function~\cite{reimers-2019-sentence-bert}. For simplicity and brevity, we refer to the fine-tuned ModernBERT model as \textit{\rmodel} in our evaluations, in contrast to the non-fine-tuned base ModernBERT~\cite{huggingfaceAlibabaNLPgtemodernbertbaseHugging, zhang2024mgte, li2023towards}. Contrastive objectives encourage the model to produce similar vector representations for semantically similar inputs (duplicate queries) while pushing apart the representations of dissimilar inputs (distinct queries). For example, ``reset my password" and ``forgot login credentials" should be close in the embedding space, whereas ``update billing info" should be far from them. However, a key limitation of conventional contrastive learning is that it treats all positive and negative examples equally. In contrast, online contrastive learning~\cite{reimers-2019-sentence-bert, sbertLossesx2014} examines an entire batch and focuses on the “hardest” examples, namely, positive pairs that the model currently ranks as relatively distant in embedding space, and negative pairs that the model ranks as relatively similar. By only computing the contrastive objective on these difficult pairs, the online contrastive loss accelerates learning in precisely the regions of the embedding space where the model is most likely to confuse positives and negatives.
This ensures that the final model is more discriminative, which is crucial for detecting subtle differences between duplicate domain-specific queries. By focusing training on challenging examples, we observe faster convergence and better precision, which is especially beneficial in semantic caching~\cite{gill2025meancacheusercentricsemanticcaching}, where correctly recognizing duplicate queries is critical for true cache hit rates.

% \waris{Self Comment: Add Synthetic Data}

\begin{figure}
  \centering
  \includegraphics[width=0.95\linewidth]{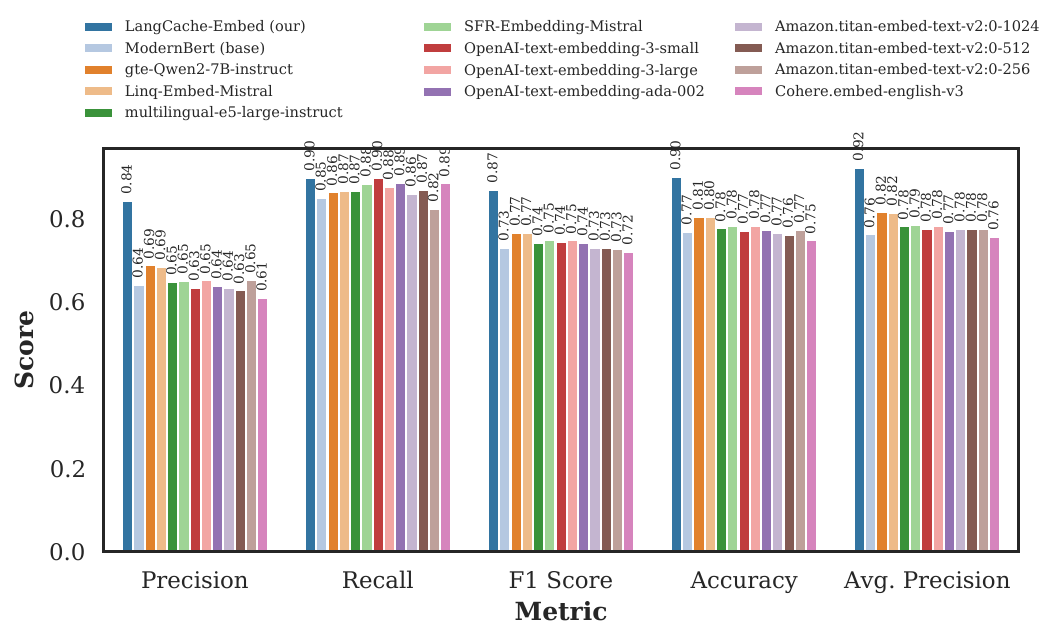}
  % \includesvg[width=0.95\linewidth, inkscapelatex=false]{svgs/comparison-train-onQuora-tested-onQuora}
  \caption{Comparison of embedding-model performance on the Quora dataset. The y-axis shows score, while the x-axis lists metrics. \rmodel (i.e., fine-tuned ModernBERT)
  exhibits a significant uplift in precision and recall compared to its baseline (non-fine-tuned) version and other state-of-the-art embedding models, highlighting the impact of fine-tuning.}
  \label{fig:ModernBERT-quora}
\end{figure}

\subsection{Synthetic Data Generation}
\label{section:Synthetic-Data-Generation}
A central challenge in developing domain-specific semantic caches is obtaining sufficient quantities of high-quality labeled data that accurately reflect the subtle ways in which users may pose similar or closely related queries. This issue is especially prominent in specialized domains such as medicine, where large and meticulously annotated datasets are often scarce. To overcome this limitation, we designed a synthetic data generation pipeline tailored to produce both positive (paraphrased) and negative (semantically related yet distinct) query pairs. This pipeline facilitates fine-tuning embedding models to more effectively distinguish near-duplicate queries from those merely related by topic.

\begin{figure}
  \centering
  \includegraphics[width=0.95\linewidth]{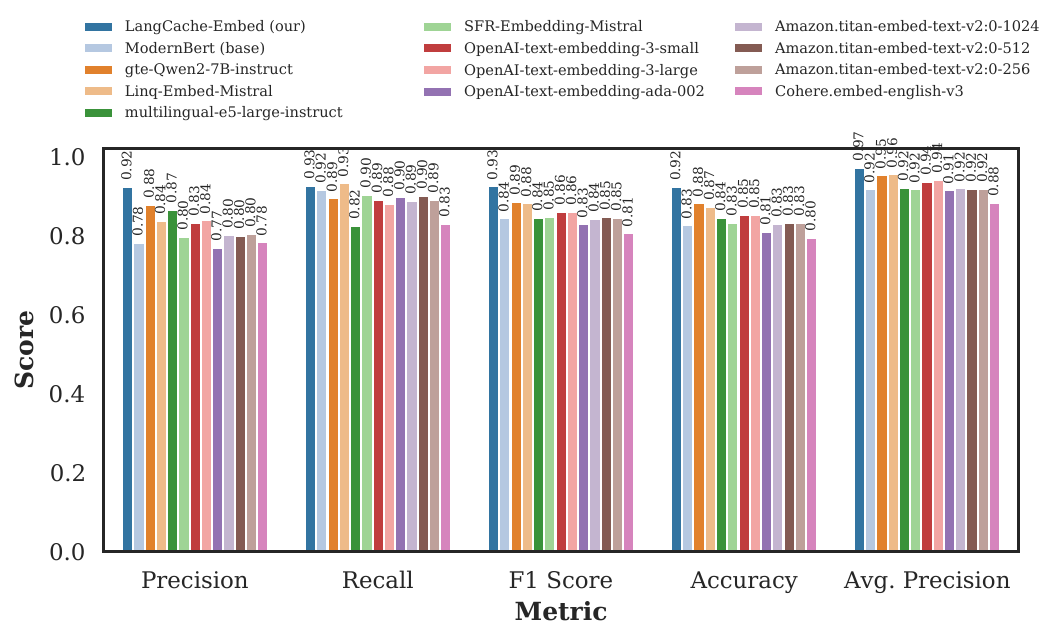}
  % \includesvg[width=0.95\linewidth, inkscapelatex=false]{svgs/comparison-train-onMedical-tested-onMedical}
  \caption{Evaluation of different embedding models on a specialized medical dataset.  Notably, \rmodel outperforms both large-scale open-source and closed-source baselines, demonstrating that lightweight models adapted to domain-specific data can achieve state-of-the-art results.}
  \label{fig:ModernBERT-medical}
\end{figure}

Our synthetic data generation approach leverages a large language model (LLM). Recognizing the frequent absence of domain-specific labeled datasets suitable for semantic caching, we capitalize instead on the widespread availability of unlabeled query datasets within the target domain (e.g., medical queries to LLMs from open-source repositories). For each original query from these datasets, carefully structured prompts guide the LLM in generating two distinct types of synthetic variants. Firstly, we create \textbf{positive samples}, which are paraphrased queries that retain the intent of the original but differ in wording or syntax. These positive samples enable the model to identify queries that, despite differences in wording, convey identical semantic meanings, thereby reducing false negatives, instances where the cache fails to recognize semantically identical queries. For example, given the unlabeled query ``Q1: What are the symptoms of early-stage diabetes?", the LLM might generate a positive sample such as ``Q2: How can I tell if someone has diabetes in its initial phase?" In this case, Q1 and Q2 are semantically identical and labeled as duplicates (i.e., a positive sample).

Secondly, we generate \textbf{negative samples}, which comprise queries that, while topically related, diverge sufficiently in focus or subdomain. These examples help embedding models identify clear semantic boundaries, thereby reducing false positives, situations in which merely related queries are incorrectly treated as near-duplicates. For instance, given the same original query ``Q1: What are the symptoms of early-stage diabetes?", a negative sample might be ``Q3: What are common health risks in children with type 1 diabetes?" Although both queries involve diabetes, Q1 focuses on general early-stage symptoms, while Q3 pertains to pediatric care and type 1 diabetes, making them semantically distinct.

Several distinctive features characterize our synthetic data generation approach. Notably, we employ a dual-labeling strategy that simultaneously produces both paraphrased ($is\_duplicate=1$) and distinct ($is\_duplicate=0$) queries within the same pipeline. This strategy effectively captures semantic edge-cases encountered in real-world applications, where user queries partially overlap but warrant distinct responses. The synthetic data generated through our pipeline plays a crucial role in fine-tuning embedding models tailored specifically for semantic caching.
%
% By exposing these models to diverse paraphrased and closely related negative examples, we enhance their capacity to detect nuanced differences in user intent, domain-specific terminologies, and clinical contexts, capabilities typically lacking in generic, pre-trained models.
By exposing these models to diverse paraphrased and closely related negative examples, we enhance their capacity to detect nuanced differences in user intent, domain-specific terminologies, and clinical contexts, which are typically lacking in generic, pre-trained models.
Consequently, this approach improves precision, by preventing confusion among semantically distinct yet related queries. Ultimately, our semantic cache achieves performance comparable to, and often surpassing, state-of-the-art closed-source and open-source embedding models on the test dataset (Section~\ref{eval:synthtetic-data}). Practically, this means organizations can attain robust, domain-specific query-handling performance without incurring the substantial computational and licensing costs typically associated with larger, resource-intensive embedding solutions.

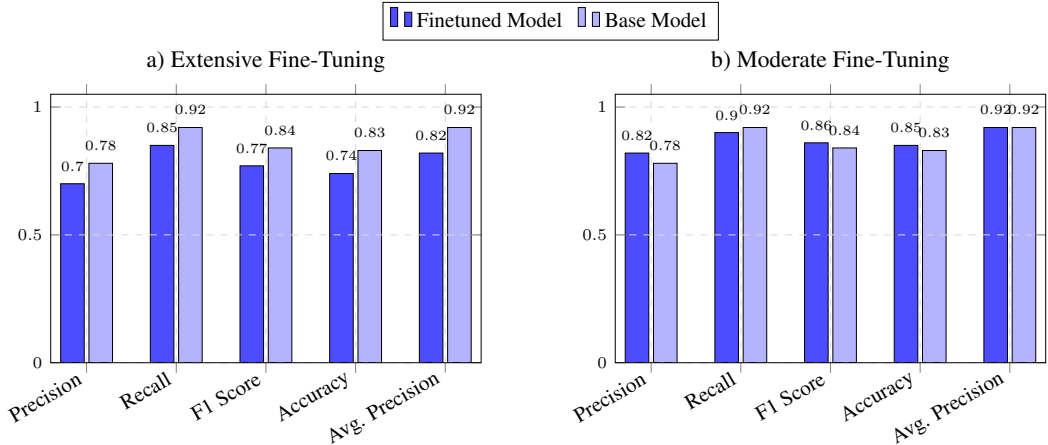
\begin{figure}
  \begin{adjustbox}{width=\textwidth}
\begin{tikzpicture}
  \begin{groupplot}[
    group style={
      group size=2 by 1,
      horizontal sep=2cm,
    },
    width=8cm,
    height=0.4\textwidth,
    ybar,
    enlarge x limits=0.1,
    ymin=0.0,
    ymax=1.05,
    symbolic x coords={Precision, Recall, F1 Score, Accuracy, Avg. Precision},
    xtick=data,
    xticklabel style={
      rotate=30,
      anchor=east,
      font=\small
    },
    nodes near coords,
    every node near coord/.append style={
      font=\tiny,
      yshift=2pt
    },
    tick label style={font=\scriptsize},
    label style={font=\small},
    axis on top,
    grid=both,
    grid style={dashed, gray!30},
    legend style={
      at={(-0.15,1.2)},
      anchor=south,
      font=\small,
      legend columns=2,
      /tikz/every even column/.append style={column sep=1ex}
    },
    legend cell align={left}
  ]

  % --- First plot ---
  \nextgroupplot[
    title={a) Extensive Fine-Tuning}
  ]

  \addplot [fill=blue!70] coordinates {
      (Precision, 0.70)
      (Recall, 0.85)
      (F1 Score, 0.77)
      (Accuracy, 0.74)
      (Avg. Precision, 0.82)
  };

  \addplot [fill=blue!30] coordinates {
      (Precision, 0.78)
      (Recall, 0.92)
      (F1 Score, 0.84)
      (Accuracy, 0.83)
      (Avg. Precision, 0.92)
  };

  % --- Second plot ---
  \nextgroupplot[
    title={b) Moderate Fine-Tuning}
  ]

  \addplot [fill=blue!70] coordinates {
    (Precision, 0.82)
    (Recall, 0.90)
    (F1 Score, 0.86)
    (Accuracy, 0.85)
    (Avg. Precision, 0.92)
  };

  \addplot [fill=blue!30] coordinates {
    (Precision, 0.78)
    (Recall, 0.92)
    (F1 Score, 0.84)
    (Accuracy, 0.83)
    (Avg. Precision, 0.92)
  };

  % Shared legend
  \legend{Finetuned Model, Base Model}
  \end{groupplot}
\end{tikzpicture}
\end{adjustbox}
\caption{The plots compare performance on the target (fine-tuning) dataset versus performance on an unseen or previously learned dataset. a) Overly extensive fine-tuning on a single domain degrades the model’s generalization to out-of-domain queries, by reduced precision. b) Limiting fine-tuning (e.g., to a single epoch and moderate gradient norm) mitigates catastrophic forgetting and preserves strong cross-domain performance.}
\label{fig:catestrophic-forgetting}
\end{figure}

\section{Evaluation}

\noindent{\textbf{Experimental Setup.}} We perform all experiments using an Amazon EC2 G6e instance~\cite{amazonAmazonInstances}, equipped with 48 parallel processes, 384 GB of system RAM, and four NVIDIA L40S GPUs, each featuring 48 GB of dedicated GPU memory. To evaluate the performance of \textit{ModernBERT}, we fine-tune the model (i.e., \rmodel) on two distinct datasets: the Quora dataset~\cite{kaggleQuestionPairs} and a specialized medical dataset~\cite{kaggleMedicalQuestion}. Each dataset comprises data points structured as pairs (Question1, Question2), accompanied by a binary label \textit{is\_duplicate}, which is set to 1 if the questions are semantically similar (i.e., duplicates), and 0 otherwise. The Quora dataset~\cite{kaggleQuestionPairs} includes $323,491$ training samples and $53,486$ evaluation samples, while the medical dataset~\cite{kaggleMedicalQuestion} consists of $2,438$ training samples and $610$ evaluation samples. The medical dataset~\cite{kaggleMedicalQuestion} introduces unique challenges, as it requires distinguishing subtle semantic variations in questions authored by 11 different medical professionals. For example, the question pair ``Can doxycycline treat an ear infection?" and ``What are the side effects of doxycycline?" is labeled 0, indicating they are not duplicates. In contrast, in the Quora dataset~\cite{kaggleQuestionPairs}, the pair ``How can I be a good geologist?" and ``What should I do to be a great geologist?" is labeled as 1, reflecting semantic duplication. Both datasets are well-suited to train specialized embeddings models for semantic caching, where the objective is to retrieve answers from cached responses for semantically similar queries, thereby reducing the need for repeated forward passes through LLMs (e.g., ChatGPT).

We compare our fine-tuned model against top-performing embedding models from the MTEB benchmark~\cite{muennighoff-etal-2023-mteb}, encompassing both open-source models including multilingual-e5-large-instruct~\cite{wang2024multilingual}, gte-Qwen2-7B-instruct~\cite{li2023towards}, Linq-Embed-Mistral~\cite{LinqAIResearch2024}, and SFR-Embedding-Mistral~\cite{SFRAIResearch2024}, as well as leading closed-source alternatives from OpenAI~\cite{Vectorem3:online}, Amazon~\cite{AmazonTi29:online}, and Cohere~\cite{EmbedThe12:online}. This comparison enables a thorough understanding of how \textit{\rmodel} performs relative to state-of-the-art solutions under different domain constraints.

% We compare our fine-tuned model against top-performing embedding models from the MTEB benchmark~\cite{muennighoff-etal-2023-mteb}, encompassing both open-source models
% multilingual-e5-large-instruct~\cite{wang2024multilingual}, gte-Qwen2-7B-instruct~\cite{li2023towards},
% Linq-Embed-Mistral~\cite{LinqAIResearch2024},
% SFR-Embedding-Mistral~\cite{SFRAIResearch2024} and leading closed-source alternatives from OpenAI~\cite{Vectorem3:online}, Amazon~\cite{AmazonTi29:online} and Cohere~\cite{EmbedThe12:online}.

We conduct the fine-tuning process using the SBERT library and employ the online contrastive loss function. Our choice of hyperparameters, one training epoch, a learning rate of \(6.5383156211679 \times 10^{-5}\), a batch size of 16, the Adam optimizer, and a gradient norm of 0.5, strikes a balance between domain-specific adaptation and the preservation of prior knowledge. Specifically, we opt for a single epoch and a relatively small gradient norm to mitigate catastrophic forgetting, as prolonged fine-tuning can diminish the model’s broader capabilities. Standard performance metrics such as Precision, Recall, F1-score, Average Precision (AP), and Accuracy are measured on both datasets, allowing us to capture the accurate comparisons.

\subsection{Domain-Specific Fine-Tuning Yields State-of-the-Art Performance}

We observe that domain-specific fine-tuning consistently enhances \textit{ModernBERT}’s performance on both Quora and medical datasets. For brevity and clear distinction, we refer to the finetuned ModernBERT as \textit{LangCache-Embed} in the evaluations. Figure~\ref{fig:ModernBERT-quora} and Figure~\ref{fig:ModernBERT-medical} illustrate the improvements in key metrics. On the Quora dataset, \textit{\rmodel}’s average precision increases from 76\% to 92\%, while on the medical dataset, it improves from 92\% to 97\%. This substantial gain highlights that fine-tuning effectively aligns model representations with domain nuances. Precision, in particular, is often critical for tasks such as semantic caching~\cite{gill2025meancacheusercentricsemanticcaching}, which require accurate retrieval of the most relevant items. Hence, we note that fine-tuning on Quora elevates \textit{\rmodel}’s precision from 64\% to 84\%, while on the medical dataset it rises from 78\% to 92\%. These consistent improvements span multiple metrics and exemplify how specialized training data can drastically refine model embeddings.

When compared to top models from the MTEB benchmark, including both open-source and proprietary options, fine-tuned \textit{ModernBERT} (i.e., \rmodel) yields state-of-the-art performance on both the Quora and medical datasets. As depicted in Figure~\ref{fig:ModernBERT-quora} and Figure~\ref{fig:ModernBERT-medical}, our model surpasses leading closed-source models and outperforms the best open-source solutions. For instance, Figure~\ref{fig:ModernBERT-medical} shows that the \rmodel achieves a 6\% improvement over OpenAI’s best embedding model (text-embedding-3-large), reinforcing that our approach maintains competitiveness even against well-established, proprietary embedding models.

\subsection{Avoiding Catastrophic Forgetting with Controlled Fine-Tuning}

Catastrophic forgetting is when a neural network forgets previously learned tasks after being trained on new ones~\cite{zhai2023investigating, franke2024preserving}. It’s a problem because it prevents models from learning continuously without losing past knowledge. This happens because neural networks update weights during training, often overwriting old information in the process. For example, if a neural network is trained to recognize animals, and then trained to recognize vehicles without revisiting the animal data, it might completely ``forget" how to recognize animals.
%
% The phenomenon of catastrophic forgetting emerges when the model is fine-tuned for multiple epochs.
%
Although extended fine-tuning can yield superior performance in a single domain, it often erodes the model’s generalization capabilities, as illustrated in Figure~\ref{fig:catestrophic-forgetting}-a.
% Over-tuning on one dataset compromises performance on out-of-domain tasks, leading to scenarios in which the non-fine-tuned base model surpasses the heavily fine-tuned variant.
%
For instance, after six epochs of fine-tuning on the Quora dataset, we observe a 22\% improvement in precision compared to the non-trained base ModernBERT on Quora test data. However, fine-tuned ModernBERT on Quora shows an 8\% drop in precision when evaluated on a medical test data, compared to its base (untrained) version, illustrating catastrophic forgetting (Figure~\ref{fig:catestrophic-forgetting}-a). This degradation indicates that excessive updates to adapt to one domain can overshadow the model’s prior knowledge. By contrast, limiting fine-tuning to a single epoch and constraining the gradient norm to 0.5 strikes a better balance in our experiments. As shown in Figure~\ref{fig:catestrophic-forgetting}-b, when we adopt these settings, the model not only retains its prior knowledge but also improves by 4\% in precision on the medical dataset after being trained on Quora dataset, thus underscoring the importance of moderate fine-tuning for preserving broader model competencies.

% \begin{figure}
%     \centering
%     \includegraphics[width=0.8\linewidth]{comparison-train-onsynthetic_medical_qwen2.5_32b-tested-onmedical.png}
%     \caption{Effect of fine-tuning ModernBERT on a purely synthetic medical dataset and then evaluating on a real-world medical dataset. The bar chart compares metrics (such as precision or average precision) before and after synthetic-data fine-tuning. Results show that leveraging synthetically generated paraphrases and related-but-different queries significantly boosts in-domain performance, allowing ModernBERT to rival or surpass larger proprietary models in terms of precision and recall.}
%     \label{fig:ModernBERT-synthetic}
% \end{figure}

\begin{table}[ht]
\centering
\small
\renewcommand{\arraystretch}{1.2}
\begin{tabular}{|lcccccc|}
\toprule
\textbf{Model} & \textbf{Source} & \textbf{Precision} & \textbf{Recall} & \textbf{F1} & \textbf{Accuracy} & \textbf{Avg. Precision} \\
\midrule
OpenAI-text-embedding-3-small & Closed & 0.83 & 0.89 & 0.86 & 0.85 & 0.94 \\
OpenAI-text-embedding-3-large & Closed & 0.85 & 0.87 & 0.86 & 0.85 & 0.94 \\
OpenAI-text-embedding-ada-002 & Closed & 0.77 & 0.90 & 0.83 & 0.81 & 0.91 \\
Amazon.titan-embed-v2:0-1024 & Closed & 0.80 & 0.89 & 0.84 & 0.83 & 0.92 \\
Amazon.titan-embed-v2:0-512  & Closed & 0.84 & 0.86 & 0.85 & 0.84 & 0.92 \\
Amazon.titan-embed-v2:0-256  & Closed & 0.80 & 0.89 & 0.85 & 0.83 & 0.92 \\
Cohere.embed-english-v3      & Closed & 0.78 & 0.83 & 0.81 & 0.80 & 0.88 \\
\midrule
Linq-Embed-Mistral              & Open & 0.84 & 0.93 & 0.88 & 0.87 & 0.96 \\
multilingual-e5-large-instruct & Open & \textbf{0.87} & 0.82 & 0.84 & 0.84 & 0.92 \\
SFR-Embedding-Mistral          & Open & 0.80 & 0.90 & 0.85 & 0.83 & 0.92 \\

gte-modernbert-base            & Open & 0.78 & 0.89 & 0.84 & 0.83 & 0.92 \\
\textbf{LangCache-Embed-Synthetic}    & Open & \textbf{0.87} & 0.90 & 0.89 & 0.88 & 0.95 \\
\bottomrule
\end{tabular}
% \caption{Performance of embedding models on the medical dataset, grouped by source.}
% \label{tab:medical_results}

\caption{Effect of fine-tuning ModernBERT on a purely \textbf{synthetic medical dataset} (\textbf{\rmodel-Synthetic}) and then evaluating on the real-world medical dataset~\cite{kaggleMedicalQuestion}. Results show that leveraging synthetic dataset significantly boosts in-domain performance, allowing ModernBERT (\textbf{\rmodel-Synthetic}) to rival or surpass larger closed-source models in both precision and recall.}
\label{table:ModernBERT-synthetic}
\end{table}

\subsection{Synthetic Data for Domain Adaptation}
\label{eval:synthtetic-data}
% \begin{tcolorbox}[mypromptbox]

\begin{lstlisting}[style=customprompt, basicstyle=\ttfamily\scriptsize,caption={Prompt used for paraphrase generation.}, label={prompt:paraphrased}]
You are a helpful medical expert. Generate 2 unique paraphrases of the given query.
Original Query: `{query}'
Each paraphrase should:
1. Preserve the original meaning but use different wording or sentence structure.
2. Avoid changing medical intent or introducing new information.
3. Be professionally written and clear.
Example:
Original Query: "What are the best ways to reduce stress?"
Paraphrased Queries:
1. ``How can a person effectively manage stress?"
2. ``What strategies help in reducing stress levels?"
Return JSON with a key `queries' containing a list of the two paraphrased versions.
\end{lstlisting}
% \end{tcolorbox}

\begin{lstlisting}[style=customprompt, basicstyle=\ttfamily\scriptsize,caption={Prompt used for non-duplicate (distinct) queries generation.}, label={prompt:non-duplicate}]
You are a helpful medical expert. Given a medical query, generate two distinct but related queries that explore different aspects of the topic.
Guidelines:
1. The new queries should be related to the original but focus on different subtopics, perspectives, or medical contexts.
2. They should not be simple rewordings or slight variations of the original.
3. Consider different patient populations, alternative diagnostic methods, treatments, or physiological explanations.
Examples:
Original Query:
"How to reduce stress?"
 Distinct Queries:
1. ``How can athletes manage stress during high-pressure competitions?" (Context: Sports Psychology)
2. ``What are effective stress management strategies for children with ADHD?" (Context: Pediatric Stress Management)
Original Query:
``A 61-year-old woman with a long history of involuntary urine loss during activities like coughing or sneezing but no leakage at night undergoes a gynecological exam and Q-tip test. Based on these findings, what would cystometry most likely reveal about her residual volume and detrusor contractions?"
Distinct Queries:
1. ``How does the Q-tip test help differentiate between stress urinary incontinence and urge incontinence?" (Context: Diagnostic Techniques)
2. ``What are the treatment options for stress urinary incontinence in postmenopausal women, and how does cystometry aid in management?" (Context: Treatment & Management)

Now, generate two distinct queries for this input:
Original Query: {query}
Return JSON with `queries' only.
\end{lstlisting}

To address \ref{prompt:paraphrased} the limited availability of annotated data in specialized domains (e.g., medical), we introduce a synthetic data generation pipeline designed to enhance domain adaptation without incurring the high costs of human annotation. Recent work by \cite{chen2024huatuogpto1medicalcomplexreasoning} has made approximately 25,000 medical queries, along with corresponding chain-of-thought reasoning and responses, publicly accessible. Leveraging this resource, we use Qwen2.5 with 32 billions parameters ~\cite{qwen2.5} to generate both semantically similar and dissimilar query pairs (Section~\ref{section:Synthetic-Data-Generation}), using prompts shown in Listings~\ref{prompt:paraphrased} and~\ref{prompt:non-duplicate}, respectively. This results in approximately 35,000 synthetic samples relevant to medical semantic task.

We fine-tune \textit{ModernBERT} (\rmodel) on this synthetic dataset and then evaluate it on a real-world medical dataset~\cite{kaggleMedicalQuestion}, observing a marked improvement. As shown in Table~\ref{table:ModernBERT-synthetic}, the precision of \textit{\rmodel} increases from 78\% to 87\%, showcasing a 9\% gain through purely synthetic data. Crucially, these results demonstrate that domain-specific knowledge can be effectively injected into the model by training on synthetic pairs that approximate realistic in-domain relationships for the purpose of embedding generation. Moreover, \rmodel now matches the performance of the best OpenAI embedding model and even exceeds it by 2\% in terms of precision, while surpassing Cohere’s closed-source embedding model by 9\% (Table~\ref{table:ModernBERT-synthetic}). This highlights the effectiveness of synthetic data generation in bridging data gaps in semantic caching, allowing models to excel in specialized domains without direct reliance on large-scale human-annotated queries, as described in Section~\ref{section:Synthetic-Data-Generation}.

% \begin{figure}
%     \input{graphs/graph2}
% \end{figure}

\subsection{Embedding Generation Overhead (Latency)}

\begin{figure}
\begin{adjustbox}{width=\textwidth}
\begin{tikzpicture}
\begin{axis}[
    width=15cm,
    height=9cm,
    xlabel={Average Embedding Generation Time},
    ylabel={Average Precision},
    legend style={at={(0.5,1.05)}, anchor=south, legend columns=3},
    grid=both,
    grid style={dashed, gray!30},
    scatter/classes={
        Finetuned-ModernBert-Quora-V1={mark=*,blue},                  
        gte-Qwen2-7B-instruct={mark=square*,red},             
        Linq-Embed-Mistral={mark=otimes,green!70!black},
        multilingual-e5-large-instruct={mark=diamond*,orange},         
        SFR-Embedding-Mistral={mark=star,purple},             
        OpenAI-text-embedding-3-small={mark=triangle,cyan},           
        OpenAI-text-embedding-3-large={mark=triangle,cyan},         
        OpenAI-text-embedding-ada-002={mark=triangle,cyan},                 
        Amazon.titan-embed-text-v2:0-1024={mark=square,teal},     
        Amazon.titan-embed-text-v2:0-512={mark=square,teal},             
        Amazon.titan-embed-text-v2:0-256={mark=square,teal},          
        Cohere.embed-english-v3={mark=+, gray}                  
    }
]

% Add shaded "most attractive quadrant for semantic cache"
\addplot [draw=none, fill=green!10, forget plot] coordinates {
    (0.0, 1)
    (0.0, 0.85)
    (0.15, 0.85)
    (0.15, 1.0)
    (0.0, 1.0)
};

% Add quadrant label
\node[align=left, anchor=north west, font=\small\bfseries, text=gray!40!black] at (axis cs:0.02,0.985) {Most attractive region\\for semantic cache.};

% Actual scatter plot
\addplot[scatter,only marks,
    scatter src=explicit symbolic, mark size=5pt]
table[meta=Model] {data.csv};

\legend{
\rmodel (our),
gte-Qwen2-7B-instruct,
Linq-Embed-Mistral,
multilingual-e5-large-instruct,
SFR-Embedding-Mistral,
OpenAI-text-embedding-3-small,
OpenAI-text-embedding-3-large,
OpenAI-text-embedding-ada-002,
Amazon.titan-embed-text-v2:0-1024,
Amazon.titan-embed-text-v2:0-512,
Amazon.titan-embed-text-v2:0-256,
Cohere.embed-english-v3
}

\end{axis}
\end{tikzpicture}
\end{adjustbox}
\caption{This plot illustrates the trade-off between embedding generation overhead (x-axis, measured in seconds) and average precision on the Quora test set (y-axis). Each point represents a different embedding model, including both open-source and commercial offerings. Models in the upper-left region deliver high precision at low embedding time. \rmodel (finetuned ModernBERT) stands out for combining rapid inference with top-tier performance, indicating it is an ideal choice for real-time semantic caching where both speed and accuracy are critical.}
\label{fig:overhead}
\end{figure}
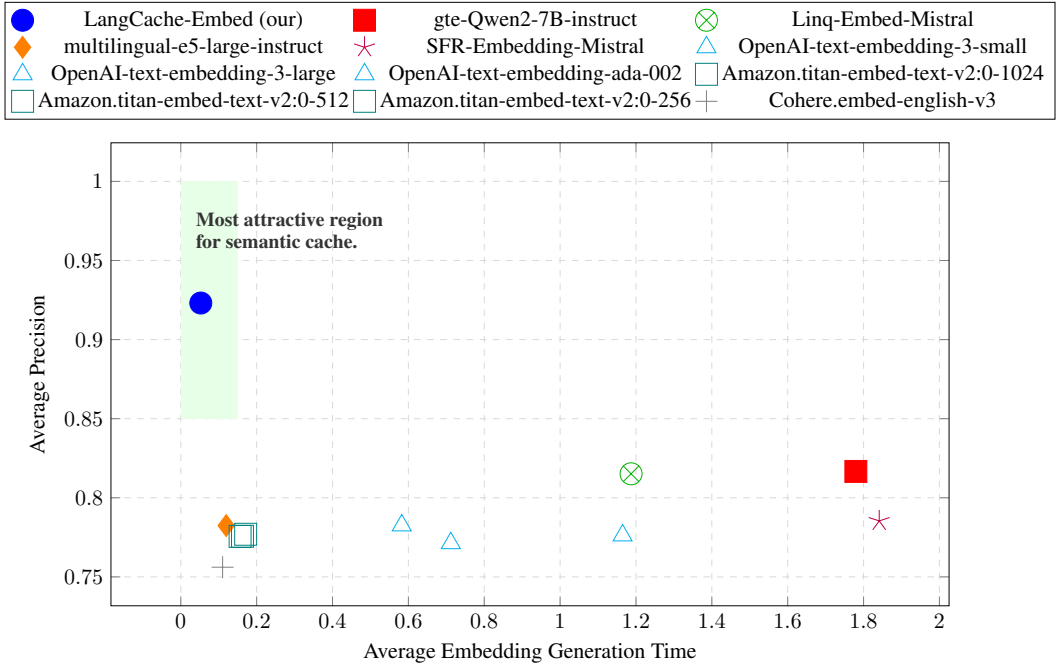

Embedding generation overhead is a critical consideration for semantic caching, as it should not disrupt the overall flow of the generative AI application. Some approaches, such as using LLMs as the embedding model (e.g., Llama) to enhance semantic caching~\cite{gptcache}, are regarded as impractical due to the high computational and memory demands~\cite{gill2025meancacheusercentricsemanticcaching}. We therefore evaluate the embedding generation overhead across a range of open-source and proprietary models, incorporating not just local computation times but also the latency of any external API calls in the case of closed-source services.

Figure~\ref{fig:overhead} illustrates the trade-off between embedding generation time (on the X-axis) and average precision on Quora test data (on the Y-axis). These measurements are obtained using CPUs rather than GPUs, as many semantic caching environments may lack access to specialized hardware. Notably, experiments on an Amazon EC2 instance reveal that Amazon Titan and Cohere exhibit comparatively lower API latencies, particularly when deployed in the same AWS region, thereby reducing network overhead.

We observe that fine-tuned \textit{ModernBERT} (\rmodel) delivers the lowest embedding generation overhead while achieving superior performance and thus emerges as an optimal choice, striking a balance between rapid embedding generation and high precision. In cases where organizational policies mandate the use of proprietary solutions, Amazon Titan also proves effective within the AWS ecosystem, offering low latency and robust performance when accessed through Bedrock APIs.

% \section{Related Work}

\section{Conclusion}
In summary, our work demonstrates that smaller embedding models, when carefully fine-tuned on domain-specific or synthetic data, can outperform significantly larger open-source and closed-source models for semantic caching. By restricting fine-tuning and carefully managing gradient norms, we avoid catastrophic forgetting and preserve general performance. Moreover, our synthetic data generation pipeline effectively addresses the scarcity of annotated domain data for semantic caching. Together, these techniques enable a lightweight, high-performing embedding model that combines efficiency with high average precision, providing a practical alternative to resource-intensive, large-scale models for semantic caching.

\bibliographystyle{plainnat}
\bibliography{main}

\end{document}